\documentclass{article}

\usepackage{microtype}
\usepackage{graphicx}
\usepackage{subcaption}
\usepackage{booktabs} \usepackage{xspace}
\usepackage{multirow}
\usepackage{graphicx}
\usepackage[table]{xcolor}
\usepackage{ifthen}
\usepackage{arydshln}
\usepackage{listings}
\usepackage{nicefrac}
\usepackage{enumitem,kantlipsum}
\def\ours{\textsc{ArgusVLM}\xspace}

\newcommand{\rank}[2][1]{\ifthenelse{\equal{#1}{1}}{\textbf{#2}}{\ifthenelse{\equal{#1}{2}}{\underline{#2}}{#2}}}

\def\onedot{.\xspace}
\def\eg{\emph{e.g}\onedot}

 \def\vs{\emph{vs}\onedot}

\usepackage{hyperref}
\usepackage{soul}
\usepackage{pifont}

\usepackage[preprint]{icml2026}

\usepackage{amsmath}
\usepackage{amssymb}
\usepackage{mathtools}
\usepackage{amsthm}

\usepackage[capitalize,noabbrev]{cleveref}

\theoremstyle{plain}

\theoremstyle{definition}

\theoremstyle{remark}

\newcommand{\circled}[1]{\tikz[baseline=(char.base)]{
    \node[draw,circle,inner sep=0.pt] (char) {\tiny\textbf{#1}};
  }}
\usepackage[textsize=tiny]{todonotes}

\icmltitlerunning{Empirical Recipes for Efficient and Compact Vision-Language Models}

\begin{document}

\twocolumn[
  \icmltitle{Empirical Recipes for Efficient and Compact Vision-Language Models}

\icmlsetsymbol{equal}{*}

  \begin{icmlauthorlist}
    \icmlauthor{Jiabo Huang}{sai,equal}
    \icmlauthor{Zhizhong Li}{sai,equal}
    \icmlauthor{Sina Sajadmanesh}{sai}
    \icmlauthor{Weiming Zhuang}{sai}
    \icmlauthor{Lingjuan Lyu}{sai}
  \end{icmlauthorlist}

  \icmlaffiliation{sai}{Sony AI}

  \icmlcorrespondingauthor{Lingjuan Lyu}{Lingjuan.Lv@sony.com}

\icmlkeywords{Vision-language Model}

  \vskip 0.3in
]

\printAffiliationsAndNotice{\icmlEqualContribution}

\begin{abstract}
Deploying vision-language models (VLMs) in resource-constrained settings demands low latency and high throughput, yet existing compact VLMs often fall short of the inference speedups their smaller parameter counts suggest.
To explain this discrepancy, we conduct an empirical end-to-end efficiency analysis and systematically profile inference to identify the dominant bottlenecks.
Based on these findings, we develop optimization recipes tailored to compact VLMs that substantially reduce latency while preserving accuracy.
These techniques cut time to first token (TTFT) by 53\% on InternVL3-2B and by 93\% on SmolVLM-256M.
Our recipes are broadly applicable across both VLM architectures and common serving frameworks, providing practical guidance for building efficient VLM systems.
Beyond efficiency, we study how to extend compact VLMs with structured perception outputs and introduce the resulting model family, \ours.
Across diverse benchmarks, \ours achieves strong performance while maintaining a compact and efficient design.
\end{abstract}

\section{Introduction}\label{sec:intro}

\begin{figure}[t]
\begin{center}
    \centerline{
      \includegraphics[width=\linewidth]{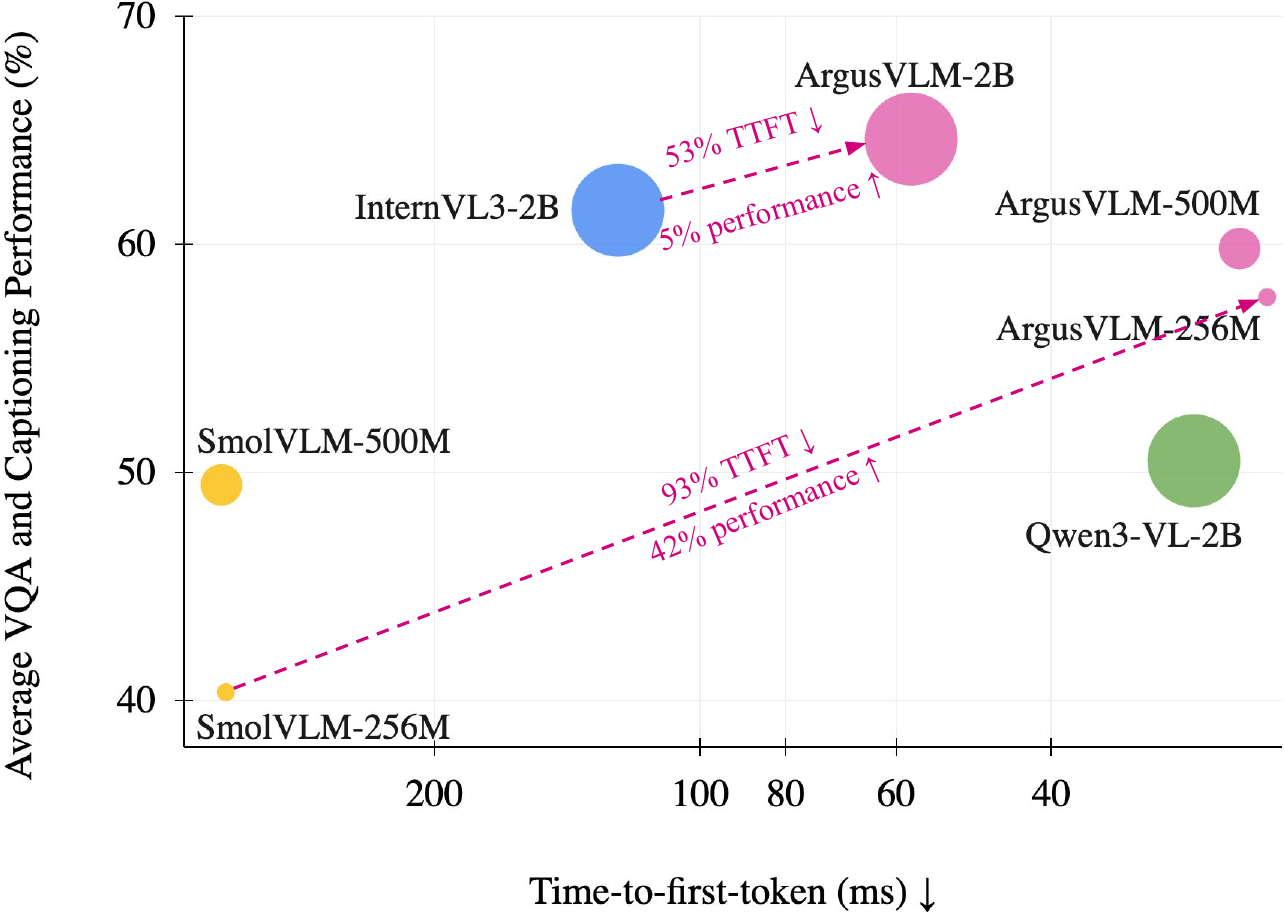}
    }
    \vspace{0.5em}
    \caption{
      \ours excels at both performance and inference efficiency with a faster time-to-first-token (TTFT) compared to existing compact VLMs.
Each bubble represents a model variant, where the area indicates model size.
    }\label{fig:teaser-efficiency}
    \vspace{-2em}
  \end{center}
\end{figure}

Building on the success of large language models, vision language models (VLMs) have advanced rapidly, enabling visual understanding through free-form, instruction-following text generation~\cite{zhu2025internvl3,bai2025qwen2}.
This progress has sparked growing interest in compact VLMs for deployment in resource-constrained settings, such as edge devices~\cite{Korrapati2024moondream,chu2023mobilevlm,chu2024mobilevlm}, due to their lower compute and memory footprint.
However, we find that these smaller models often do not deliver the expected end-to-end efficiency gains.
For instance, despite having far fewer parameters, SmolVLM-256M~\cite{marafioti2025smolvlm} has a longer time-to-first-token (TTFT; 344.7\,ms) than InternVL3-8B~\cite{zhu2025internvl3} (177.4\,ms).

\begin{figure}[t]
\begin{center}
    \centerline{
      \includegraphics[width=\linewidth]{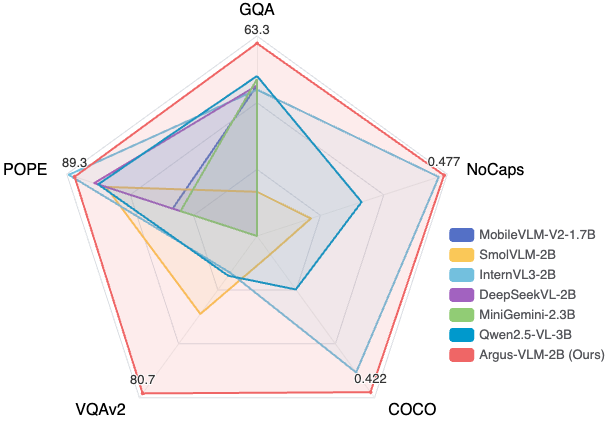}
    }
    \vspace{0.5em}
    \caption{
      \ours-2B achieves strong performance on image understanding and captioning tasks across five benchmarks compared with leading vision-language models such as QwenVL~\cite{bai2025qwen2} and InternVL~\cite{zhu2025internvl3}.
    }\label{fig:teaser-performance}
  \end{center}
  \vspace{-1em}
\end{figure}

In this work, we conduct a comprehensive study of compact VLMs with a focus on their end-to-end efficiency.
Through systematic profiling, we find that in the compact regime, previously overlooked CPU-side operations, such as image processing and text tokenization, often dominate latency.
In contrast, GPU-side computations contribute less to overall latency, due to extensive prior optimizations such as FlashAttention~\cite{dao2022flashattention}, CUDA graphs~\cite{nvidia_cuda_graphs}, and kernel fusion~\cite{filipovivc2015optimizing}.
Motivated by these findings, we develop targeted optimizations that substantially reduce inference latency without compromising accuracy.
As a result, we reduce TTFT by 53\% on InternVL3-2B (\(124.0 \rightarrow 57.7\)\,ms) and by 93\% on SmolVLM-256M (\(344.7 \rightarrow 22.8\)\,ms), as shown in \cref{fig:teaser-efficiency}.

Beyond inference efficiency, we investigate how compact VLMs can be extended from multimodal understanding to visual perception tasks that require structured, region-level outputs.
We focus on dense image captioning~\cite{johnson2016densecap}, a practical and challenging task that pairs localized image regions with semantically rich captions.
While prior VLM approaches generate bounding boxes either as numeric coordinates in text or via specialized location tokens, it remains unclear which design is more effective, especially in the compact regime.
To clarify this, we conduct a controlled comparison of these two formulations and identify the most suitable approach for enabling compact VLMs to perform structured prediction under a language-modeling objective.

Our contributions are threefold.
\textbf{(i)} We present a systematic end-to-end profiling study of compact VLMs, pinpointing the primary latency bottlenecks and deriving actionable guidance for practical deployment.
\textbf{(ii)} We introduce \ours, a family of compact VLMs spanning 256M--2B parameters, and conduct a controlled study of bounding-box prediction formulations to better unify structured visual perception with text-based multimodal understanding.
\textbf{(iii)} \ours achieves strong performance on diverse VQA and captioning benchmarks (\cref{fig:teaser-performance}), as well as dense image captioning,
delivering up to 93\% and 80\% reductions in TTFT and end-to-end latency, respectively, compared to SmolVLM-256M.
 \section{Related Works}\label{sec:related}

\subsection{Vision-language Models}

Vision-language models (VLMs) have rapidly evolved over the past few years.
Pioneering methods such as CLIP~\cite{radford2021clip,zhai2023siglip,sun2023eva}
adopt a dual-encoder design to learn discriminative image representations by aligning image and text embeddings using language supervision.
Inspired by the success of generative language modeling~\cite{radford2018improving,brown2020language},
later works adopt an encoder-decoder architecture~\cite{raffel2020exploring} that unifies diverse computer vision tasks as a token generation problem with an end-to-end training~\cite{lu2022unified,lu2024unified,xiao2024florence}.
Following the success of LLaVA~\cite{liu2023visual} on visual instruction tuning,
recent developments~\cite{li2024llava,chen2024expanding,zhu2025internvl3,bai2025qwen2,team2025kimi} have gradually converged to the decoder-only architecture,
where visual embeddings from a pretrained vision encoder are used as conditions for text generation via a pre-trained large language model (LLM)~\cite{yang2025qwen3,cai2024internlm2,allal2025smollm2}.
This design greatly improves flexibility and generalization across tasks.

\subsection{Compact Vision-Language Models}

There is a growing interest in building compact VLMs, usually with fewer than 2 billion parameters, that can deliver strong multimodal performance under limited resources.
Moondream~\cite{Korrapati2024moondream} emphasizes both efficiency and performance but remains closed-source.
H2OVL~\cite{galib2024h2ovl} explores compact VLMs optimized for OCR-related tasks, yet its capability to generalize to broader visual domains is underexplored.
SmolVLM~\cite{marafioti2025smolvlm} highlights the importance of memory footprint by an extensive empirical study of architectural trade-offs for compact models.
However, it fails to achieve the expected inference speed up compared to larger models and are primarily evaluated on language-based understanding tasks, overlooking structured perception problems such as dense captioning.

\subsection{Efficiency of VLMs}
Beyond reducing model size, recent VLMs improve runtime efficiency through techniques such as visual token compression~\cite{zhu2025internvl3,marafioti2025smolvlm,chu2023mobilevlm,chu2024mobilevlm} and more efficient multimodal fusion~\cite{zhang2025llava}, which alleviate the computation overhead induced by long visual token sequences.
In parallel, serving systems such as vLLM~\cite{kwon2023efficient} provide practical, general-purpose optimizations (\eg, kernel-level speedups and scheduling) that significantly improve throughput and latency.
Our methods target end-to-end bottlenecks and identify overlooked bottlenecks on top of generic serving frameworks.
We observe additional inference speed improvements even when deploying models with vLLM.

\subsection{Perception Support of VLMs}
Visual perception capabilities are increasingly recognized as essential and have been integrated into several state-of-the-art VLMs.
Most existing approaches represent object locations either as plain-text numerical coordinates~\cite{zhu2025internvl3,bai2025qwen2} or via dedicated special tokens~\cite{xiao2024florence,lu2022unified}.
However, it remains unclear which formulation is more effective, particularly for compact VLMs, where limited model capacity may amplify representational bottlenecks.
In this work, we provide a controlled comparison of these design choices and study how to best equip compact VLMs with structured perception outputs under a language-modeling objective.
 \section{Recipes to Improve Inference Efficiency}\label{sec:method-efficiency}

To improve the inference efficiency of compact VLMs, we build upon the efficient serving framework vLLM~\cite{kwon2023efficient}.
vLLM is a highly optimized system for serving large language models with low latency and state-of-the-art throughput~\cite{vllm_docs}.

\subsection{Inference Computation Profiling}\label{sec:profiling}

Accurately identifying bottlenecks is crucial for improving VLM efficiency and requires reliable profilers.
We therefore evaluated several profilers to measure the time of both CPU- and GPU-side operations in end-to-end inference.

\begin{figure}[t]
  \centering
  \includegraphics[width=\linewidth]{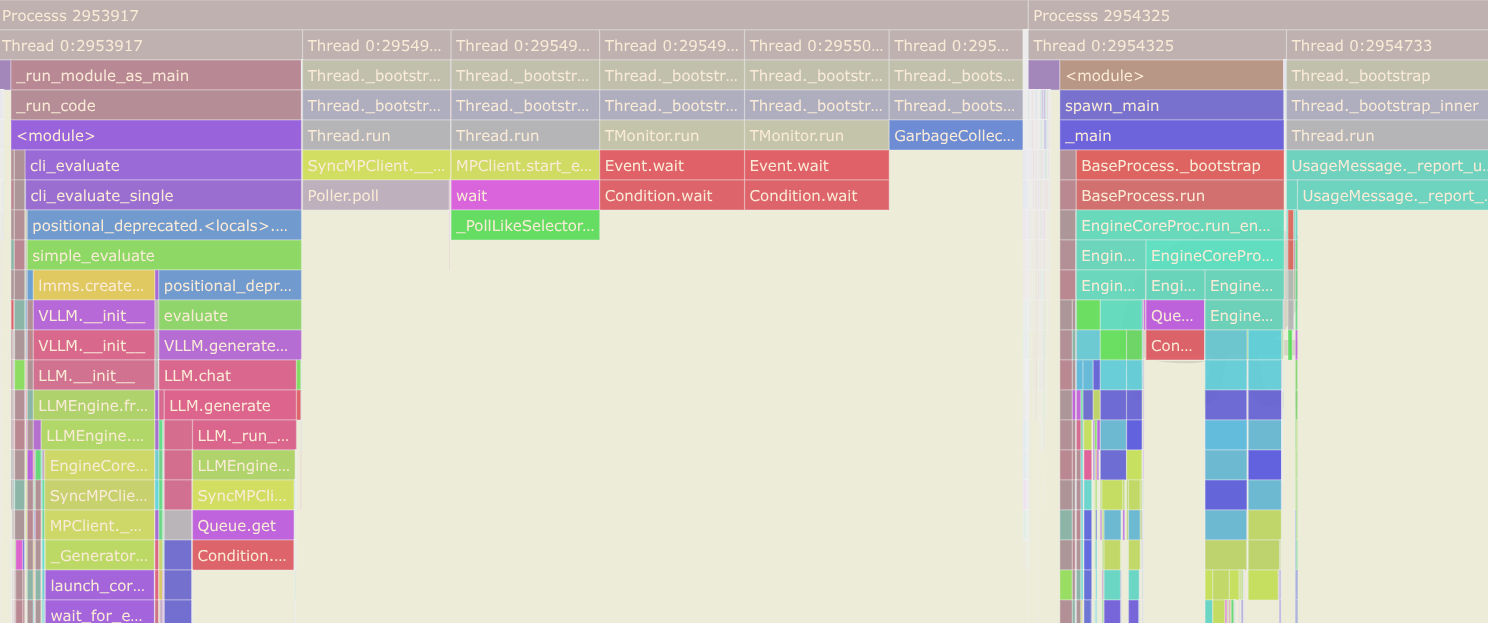}
  \caption{
    Flame graph from profiling VLM inference with \emph{austin} \cite{austin} to identify CPU-side bottlenecks.
    The visualization separates processes for easier analysis: Process 2953917 performs multimodal preprocessing, while Process 2954325 runs model inference.
    Best viewed digitally with zoom.
  }\label{fig:austin}
  \vspace{-0.em}
\end{figure}

\paragraph{Profilers.}

We choose \textit{austin} \cite{austin} for CPU-side profiling and NVIDIA Nsight Systems \cite{nsight} after evaluating the following profilers:
\vspace{-0.5em}
\begin{itemize}[leftmargin=0.4cm]
  \itemsep0em
  \item \textbf{cProfile}~\cite{python2025profile} is a built-in, deterministic Python profiler that provides function-level measurements and call counts.
        However, it incurs higher overhead than sampling profilers and offers limited support for visualizing complex codebases such as vLLM.
  \item \textbf{py-spy}~\cite{pyspy} and \textbf{austin}~\cite{austin} are low-overhead sampling profilers that can attach to a running Python process without code changes and provide line-level attribution.
        Both can generate flame graphs to surface hotspots.
        Compared to \emph{py-spy}, \emph{austin} better supports multi-process applications, which is important for profiling vLLM because it uses multiple processes to handle requests (see \cref{fig:austin}).
        We therefore recommend \emph{austin} for CPU-side profiling.
\item \textbf{NVIDIA Nsight Systems}~\cite{nsight} provides a system-level view of GPU activity, including kernel timings, memory transfers, and CPU--GPU interactions.
        We use it to diagnose GPU-side bottlenecks, particularly to understand why quantization may not yield speedups for compact VLMs.
\end{itemize}

\begin{figure}[t]
  \centering
  \begin{subfigure}[b]{0.99\linewidth}
    \includegraphics[width=\linewidth]{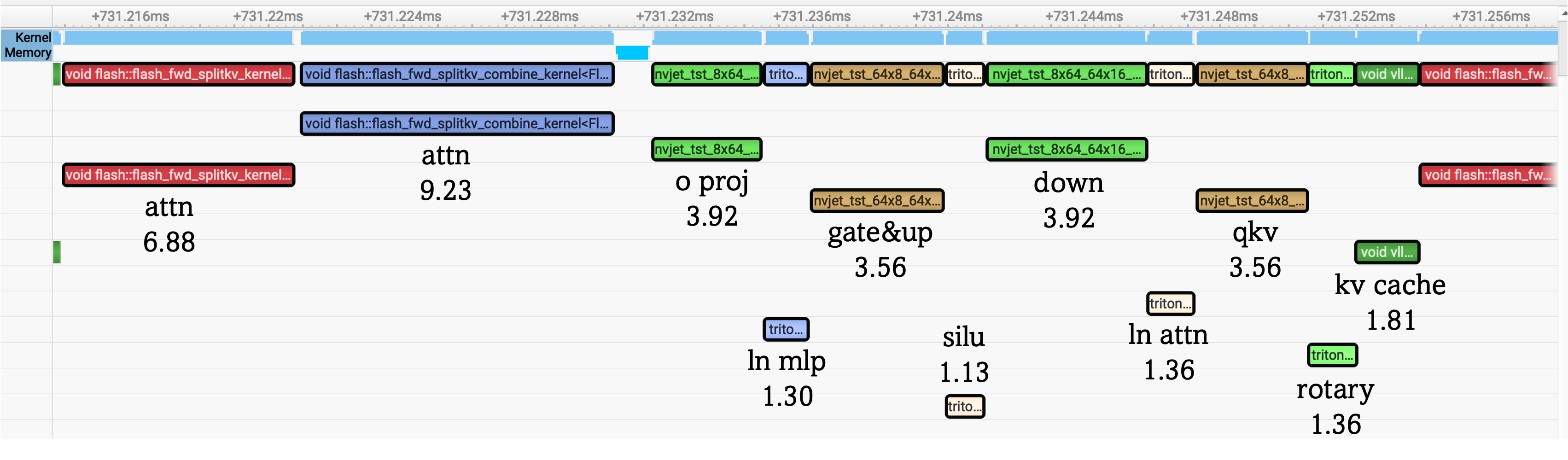}
    \caption{SmolVLM-256M with bfloat16 precision.
    }\label{fig:smol_nsys}
  \end{subfigure}
  \begin{subfigure}[b]{0.99\linewidth}
    \includegraphics[width=\linewidth]{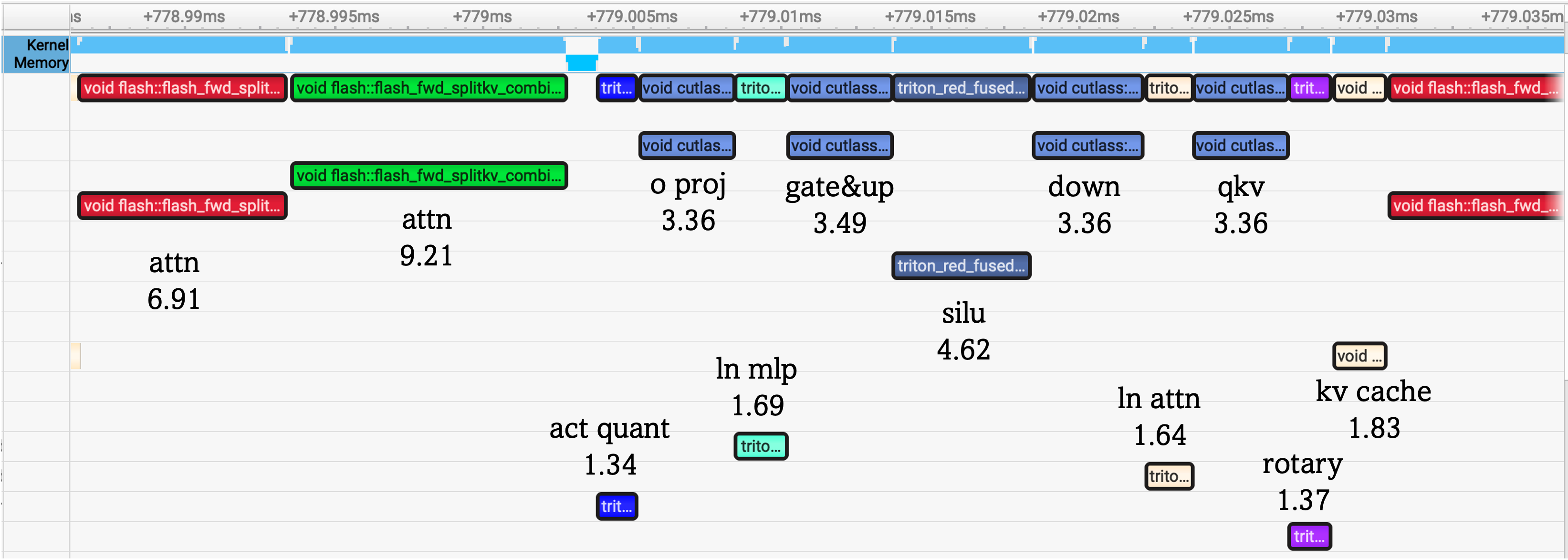}
    \caption{SmolVLM-256M with W8A8 quantization.
    }\label{fig:smol_quant_nsys}
  \end{subfigure}
  \vspace{-0.5em}
  \caption{
    GPU profiling of SmolVLM-256M inference under vLLM with (a) bfloat16 and (b) W8A8 quantization.
    The timelines show GPU kernels within a transformer layer during decoding, with the average execution time (\(\mu\)s) annotated below each kernel. Despite reduced precision, the quantized model is slower.
  }\label{fig:nsys}
  \vspace{-1.2em}
\end{figure}
\paragraph{Quantization Remains Ineffective for Compact VLMs.}

We quantize our \ours-256M (introduced in \cref{sec:method-vlm}) using \emph{llmcompressor}~\cite{llmcompressor2024} with FP8 W8A8 quantization, and evaluate inference efficiency under vLLM against the bfloat16 baseline.
Surprisingly, the quantized model achieves 6.4\% lower decoding throughput than bfloat16 (499.08 \vs 533.28 tokens/s on an H100 GPU).
To identify the cause, we profile both settings and inspect the GPU timelines for one transformer layer during decoding (\cref{fig:nsys}).
Quantized matrix multiplications, including QKV projection, output projection, gate-up projection, and down projection, are indeed faster (-1.39\(\mu\)s) by \(1.1\times\).
However, activation quantization introduces substantial overhead (+5.50\(\mu\)s), despite being fused into the \emph{layernorm} and \emph{silu} kernels.
Overall, this overhead outweighs the gains from quantized matrix multiplications, leading to lower throughput.
These results suggest that fully FP8 execution (avoiding bfloat16\(\rightarrow\)FP8 conversions) is a promising direction for improving quantized inference in compact VLMs.
\vspace{-0.5em}
\paragraph{CPU-Side Operations are the Primary Bottleneck.}
Profiling in \cref{fig:nsys} also highlights the extent of existing GPU optimizations in vLLM:
(1) compute-intensive operators leverage optimized kernels such as FlashAttention~\cite{dao2022flashattention};
(2) kernels are fused where possible (\eg, \emph{layernorm} and \emph{rotary embedding}) to reduce memory traffic;
and (3) CUDA Graphs reduce kernel-launch overhead, resulting in near-continuous kernel execution.
Together, these optimizations push GPU-side performance close to saturation.
As a result, end-to-end latency in compact VLMs is often dominated by CPU-side workload, which is frequently overlooked in prior optimization efforts.
We therefore focus on CPU-side bottlenecks, which directly impact TTFT -- a key metric for interactive user experience.

\subsection{Case Studies: InternVL3-2B and SmolVLM-256M}\label{sec:case-study}

We next present two case studies---InternVL3-2B~\cite{zhu2025internvl3} and SmolVLM-256M~\cite{marafioti2025smolvlm}---to illustrate common implementation inefficiencies in compact VLM inference and the impact of targeted fixes.

\vspace{-.5em}
\paragraph{Settings.}
We use LMMs-Eval~\cite{zhang2024lmmsevalrealitycheckevaluation} to run the COCO2017~\cite{lin2014microsoft} image captioning benchmark with vLLM as the serving backend.
The benchmark contains 5{,}000 multimodal requests, evaluated with a batch size of 1.
During inference, we profile CPU-side execution using \emph{austin} and inspect flame graphs (\eg, \cref{fig:austin}) to locate hotspots.
We then apply targeted optimizations, re-measure latency, and iterate until the remaining costs are no longer dominated by obvious implementation bottlenecks.

\begin{table}[t]
  \centering
  \small
  \caption{Impact of incremental optimizations on InternVL3-2B inference.
    The first row is baseline performance under vLLM.
}\label{tab:internvl_optim}
  \vspace{-0.3em}
  \setlength{\tabcolsep}{1pt}
  \begin{tabular}{p{0.46\linewidth}ccc}
    \toprule
    \multirow{2}{*}{Optimizations}                        & TTFT                   & Throughput              & E2E Latency             \\
                                                          & (ms)$\downarrow$       & (tokens/s)$\uparrow$    & (ms)$\downarrow$        \\
    \midrule
    InternVL3-2B                                          & 124.0                  & 270.86                  & 180.85                  \\
    ~~ + Reduce img transform \ding{172}\,\&              & \multirow{3}{*}{~80.9} & \multirow{3}{*}{279.14} & \multirow{3}{*}{127.20} \\
    ~~~\,\,\,~Tensor img process \ding{173}  \&                                                                                        \\
    ~~~\,\,\,~GPU preprocess \ding{174}                                                                                                \\
    ~~ + Pin memory  \ding{175}                           & ~72.8                  & 278.43                  & 119.31                  \\
    ~~ + Reduce PIL decoding  \ding{176}                  & ~71.7                  & 280.30                  & 109.52                  \\
    ~~ + BF16 img normalize  \ding{177}                   & ~68.4                  & 282.87                  & 113.40                  \\
    ~~ + Tokenizer  \ding{178} \& UInt8  \ding{179}       & ~58.3                  & 281.94                  & 105.82                  \\
    ~~ + Pack H2D transfer \raisebox{0.1ex}{\circled{12}} & ~\textbf{57.7}         & \textbf{292.66}         & \textbf{101.26}         \\
    \bottomrule
  \end{tabular}
  \vspace{-.5em}
\end{table}

\vspace{-.5em}
\paragraph{InternVL3-2B.}
As summarized in \cref{tab:internvl_optim}, the optimizations reduce TTFT from \(124.0\) to \(57.7\)\,ms (-53\%).
End-to-end latency (from request arrival to the full response) drops from \(180.85\) to \(101.26\)\,ms (-44\%).

\vspace{-.5em}
\paragraph{SmolVLM-256M.}
We apply the same workflow to SmolVLM-256M and report results in \cref{tab:smolvlm_optim}.
TTFT decreases from \(344.7\) to \(22.8\)\,ms (-93\%), and end-to-end latency descends from \(427.59\) to \(85.07\)\,ms (-80\%).
\begin{table}[t]
  \centering
  \small
  \caption{Impact of incremental optimizations on SmolVLM-256M inference.
    The first row is baseline performance under vLLM.
}\label{tab:smolvlm_optim}
  \vspace{-0.3em}
  \setlength{\tabcolsep}{1pt}
  \begin{tabular}{p{0.46\linewidth}ccc}
    \toprule
    \multirow{2}{*}{Optimizations}                         & TTFT                   & Throughput              & E2E\,Latency            \\
                                                           & (ms)$\downarrow$       & (tokens/s)$\uparrow$    & (ms)$\downarrow$        \\
    \midrule
    SmolVLM-256M                                           & 344.7                  & 460.71                  & 427.59                  \\
    ~~\,+ Pillow-SIMD  \ding{180}                          & 241.6                  & 461.12                  & 325.20                  \\
    ~~~+ Reduce img transform\,\ding{172}\,\&              & \multirow{3}{*}{~71.9} & \multirow{3}{*}{420.91} & \multirow{3}{*}{151.34} \\
    ~~~\,\,\,~Tensor img process \ding{173}  \&                                                                                         \\
    ~~~\,\,\,~Reduce PIL decoding  \ding{176}                                                                                           \\
    ~~ + GPU preprocess  \ding{174}                        & ~59.7                  & 466.21                  & 131.65                  \\
    ~~ + UInt8  \ding{179}                                 & ~47.4                  & 471.77                  & 118.32                  \\
    ~~ + BF16 img normalize  \ding{177}                    & ~38.3                  & 470.40                  & 109.71                  \\
    ~~ + Remove pixel mask  \raisebox{0.1ex}{\circled{10}} & ~29.5                  & 494.51                  & ~~96.54                 \\
    ~~ + Pin memory   \ding{175}                           & ~28.7                  & 497.95                  & ~~95.26                 \\
    ~~ + Tokenizer   \ding{178}                            & ~27.4                  & 498.86                  & ~~93.91                 \\
    ~~ + Avoid split \raisebox{0.1ex}{\circled{11}}        & ~23.5                  & 500.74                  & ~~89.67                 \\
    ~~ + Pack H2D transfer \raisebox{0.1ex}{\circled{12}}  & ~\textbf{22.8}         & \textbf{533.28}         & \textbf{~~85.07}        \\
    \bottomrule
  \end{tabular}
  \vspace{-0.5em}
\end{table}

\subsection{Practical Recipes to Reduce Latency}\label{sec:recipes}

Based on these case studies, we distill the following practical recipes for reducing latencies in compact VLMs:
\vspace{-0.25em}
\begin{itemize}[leftmargin=0.4cm]
  \itemsep0em
  \item \textbf{Minimize image processing on the critical path.}
        (a) Avoid repeated image decode/encode\,\ding{176}, resizing, cropping and padding\,\ding{172}.
        (b) Use optimized image libraries \eg, \emph{Pillow-SIMD}~\cite{pillowsimd} for decoding\,\ding{180}, torchvision for transforms\,\ding{173}.
        (c) When possible, offload preprocessing (resize/normalize) to GPU\,\ding{174}.
        (d) Simplify the image processing logic wherever possible\,\ding{177}\,\raisebox{0.4ex}{\circled{10}}\,\raisebox{0.4ex}{\circled{11}}.
  \item \textbf{Reduce CPU--GPU communication.}
        (a) Transfer smaller dtypes (\eg, UInt8 images instead of Float32) and convert on-device\,\ding{179}.
        (b) Use pinned (page-locked) memory to accelerate H2D (host-to-device) transfers\,\ding{175}.
        (c) Packing small H2D transfers into one transfer\,\raisebox{0.4ex}{\circled{12}}.
  \item \textbf{Tokenization can be slow.}
        The large number of repeated image token placeholders (\eg,\,\texttt{\textless IMG\_CONTEXT\textgreater}) in VLMs leads to extremely long prompts, which can slow down tokenization.
        Using a more compact placeholder scheme yields measurable TTFT gains\,\ding{178}.
  \item \textbf{Profile first, optimize later.}
        Use profilers such as \emph{austin} and \emph{Nsight Systems} to identify real bottlenecks before optimization and avoid premature optimizations.
\end{itemize}

 \section{Recipes for Building Compact VLMs}\label{sec:method-vlm}

\begin{figure*}[t]
\centering
\includegraphics[width=0.99\textwidth]{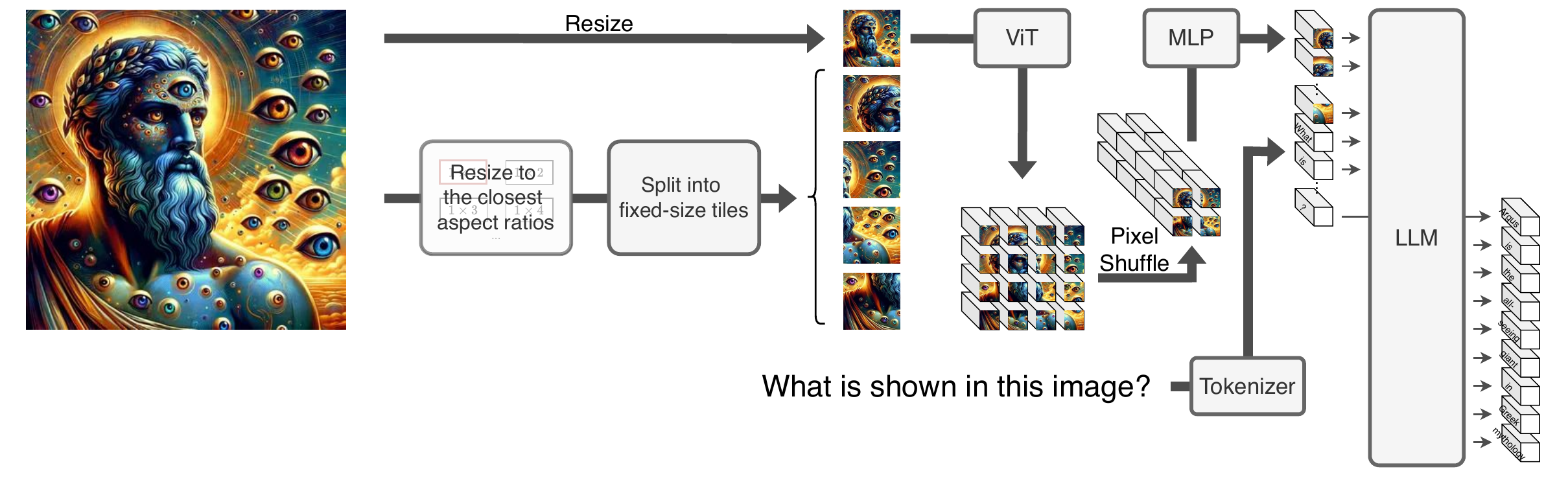}
\vspace{-0.2em}
\caption{Overview of the \ours architecture.
\ours consists of a Vision Transformer (ViT), a single-layer MLP, and a large language model (LLM).
Given an input image, we first resize it to the closest aspect ratio from a predefined set, then split it into square tiles that are resized to the ViT input resolution; we also generate a global thumbnail to provide scene-level context.
Each tile (and the thumbnail) is independently encoded by the ViT into patch tokens, which are then compressed by concatenating features from spatially adjacent patches along the channel dimension.
In parallel, the user instruction is tokenized into text tokens.
Finally, the visual and text tokens are concatenated and fed to the LLM, which generates the response autoregressively.
}\label{fig:architecture}
\vspace{-0.5em}
\end{figure*}

Beyond efficiency, a practical challenge for VLMs is adapting them to downstream tasks that are not explicitly covered during pre-training.
For example, although several recent models support visual grounding~\cite{zhu2025internvl3,qwen2024qwen2_5}, few can perform dense captioning, a useful capability that requires structured, region-level outputs.
To best enable dense captioning in the compact regime, we conduct a controlled comparison of two common formulations for learning regional labels under a next-token prediction objective.
Based on the most effective design, we introduce \ours, a family of compact VLMs that supports dense captioning while maintaining strong performance on conventional text-based image understanding tasks.
In this section, we first describe the \ours architecture and then detail the two bounding-box prediction formulations.

\subsection{Model Architecture}
\label{sec:arch}
We follow recent decoder-only VLMs~\cite{marafioti2025smolvlm,zhu2025internvl3} and adopt the ``ViT--MLP--LLM'' architecture~\cite{chen2024internvl} shown in \cref{fig:architecture}.
A pretrained Vision Transformer (ViT)~\cite{vaswani2017attention} encodes an input image into a sequence of visual embeddings.
In parallel, the textual instruction (or question) is tokenized into text tokens and embedded using the vocabulary of a pretrained large language model (LLM).
An MLP then projects the visual embeddings into the LLM embedding space to align feature dimensions across modalities.
Finally, the projected visual tokens are concatenated with the text tokens and passed to the LLM for autoregressive generation.

\paragraph{Image Encoding.}
Inspired by UReader~\cite{ye2023ureader}, we adopt an image-tiling strategy to preserve fine-grained visual details.
Most off-the-shelf ViTs are pretrained on fixed-resolution square images, so na\"ively resizing high-resolution inputs with arbitrary aspect ratios often introduces distortion and obscures important visual details.
To mitigate this issue, we first resize each image to the closest aspect ratio from a predefined set (minimizing geometric distortion), then split it into square tiles and resize each tile to the ViT's native input resolution.
In addition, we create a global thumbnail to provide scene-level context.
All tiles and the thumbnail are encoded independently by the vision encoder, and their resulting visual token sequences are concatenated to form the visual input to the LLM.
This design enables robust handling of diverse aspect ratios while retaining both local detail and global context.

\paragraph{Visual Token Reduction.}
Image tiling preserves fine-grained details but increases the number of visual tokens, which can slow down the LLM.
To reduce this overhead, \ours applies pixel-unshuffle (space-to-depth)~\cite{shi2016real,marafioti2025smolvlm} to the ViT patch grid.
Specifically, we merge each $r \times r$ neighborhood of spatially adjacent patch tokens into a single token by concatenating their features along the channel dimension.
If the original visual token sequence has shape $(N, D)$, this produces a compressed sequence of shape $(\nicefrac{N}{r^2}, D\cdot r^2)$, yielding an $r^2$ reduction in token count while largely preserving local visual information.
This design provides a favorable trade-off between visual fidelity and inference efficiency.

\paragraph{Training and Inference.}
We train \ours with teacher forcing~\cite{williams1989learning}, predicting each output token conditioned on the preceding ground-truth tokens.
All training samples are reformatted into dialogues between a ``User'' and an ``Assistant'' using a unified chat template.
Given an input image and instruction, we tokenize the dialogue and concatenate the resulting text tokens with the corresponding visual tokens to form an input sequence
$X \in \mathbb{R}^{T \times d}$,
with sequence length $T$ and hidden dimension $d$.
We construct supervision by shifting the sequence by one position so that the model learns next-token prediction.
We optimize a categorical cross-entropy loss computed only over the assistant response tokens:
\begin{equation}
\mathcal{L}(\theta) = -\sum_{i=t}^{T} \log P_\theta(x_i \mid x_{<i}),
\label{eq:ntp}
\end{equation}
where $\theta$ denotes the model parameters, $t$ is the index of the first assistant token, and $P_\theta(x_i \mid x_{<i})$ is the predicted probability of the ground-truth token $x_i$ given all previous tokens.
During inference, we apply the same chat template to the user's instruction and image(s) and leave the assistant response empty for autoregressive generation.

\begin{table*}[t]
    \centering
    \caption{Comparison to the leading vision-language models on text-based image understanding benchmarks. Results of models marked with * are measured by their open-source checkpoints provided on Huggingface~\cite{wolf2019huggingface} using LMMs-Eval~\cite{zhang2024lmmsevalrealitycheckevaluation}.
    }\label{tab:sota}
\resizebox{\textwidth}{!}{\begin{tabular}{@{}lcccccccc@{}}\toprule
            \multirow{4}{*}{\textbf{Model}}                & \multirow{4}{*}{\textbf{\#Parameters}} & \multicolumn{3}{c}{\textbf{VQA}} & \multicolumn{4}{c}{\textbf{Captioning}}                                                                                                                           \\\cmidrule{3-9}
                                                           &                                        & \textbf{GQA}                     & \textbf{POPE}                           & \textbf{VQAv2} & \multicolumn{2}{c}{\textbf{COCO2017}} & \multicolumn{2}{c}{\textbf{NoCaps}}                            \\\cmidrule{3-9}
                                                           &                                        & Exact Match ↑                    & Accuracy ↑                              & Accuracy ↑     & BLEU4 ↑                               & CIDEr ↑                             & BLEU4 ↑        & CIDEr ↑ \\\midrule
            SmolVLM*~\cite{marafioti2025smolvlm}           & 256M                                   & 41.5                             & 80.1                                    & 66.0           & 0.056                                 & 0.118                               & 0.087          & 0.182   \\
            \rowcolor{blue!5} \ours (Ours)                 & 256M                                   & 55.3                             & 87.3                                    & 70.4           & 0.347                                 & 1.191                               & 0.409          & 1.014   \\\midrule
            SmolVLM*~\cite{marafioti2025smolvlm}           & 500M                                   & 45.3                             & 85.9                                    & 70.4           & 0.184                                 & 0.712                               & 0.273          & 0.722   \\
            \rowcolor{blue!5} \ours (Ours)                 & 500M                                   & 58.2                             & 86.3                                    & 74.1           & 0.369                                 & 1.263                               & 0.438          & 1.086   \\\midrule
            MobileVLM~\cite{chu2023mobilevlm}              & 1.7B                                   & 56.1                             & 84.5                                    & -              & -                                     & -                                   & -              & -       \\
            MobileVLM-V2~\cite{chu2024mobilevlm}           & 1.7B                                   & 59.3                             & 84.3                                    & -              & -                                     & -                                   & -              & -       \\
            SmolVLM*~\cite{marafioti2025smolvlm}           & 2B                                     & 49.2                             & 87.7                                    & 75.3           & 0.226                                 & 0.889                               & 0.351          & 0.932   \\
            InternVL3*~\cite{zhu2025internvl3}             & 2B                                     & 58.9                             & \textbf{89.6}                           & 72.5           & 0.394                                 & 1.371                               & 0.472          & 1.208   \\
            DeepSeekVL~\cite{lu2024deepseekvl}             & 2B                                     & 59.3                             & 88.3                                    & -              & -                                     & -                                   & -              & -       \\
            MiniGemini~\cite{li2024minigemini}             & 2.3B                                   & 59.9                             & 83.9                                    & -              & -                                     & -                                   & -              & -       \\
            Qwen2.5-VL*~\cite{bai2025qwen2}                & 3B                                     & 60.2                             & 88.1                                    & 72.7           & 0.276                                 & 1.014                               & 0.399          & 1.047   \\
            Qwen3-VL*~\cite{bai2025qwen3vltechnicalreport} & 2B                                     & 59.4                             & 89.4                                    & 74.1           & 0.127                                 & 0.220                               & 0.171          & 0.194   \\
            \rowcolor{blue!5} \ours (Ours)                 & 2B                                     & \textbf{63.3}                    & 89.3                                    & \textbf{80.7}  & \textbf{0.422}                        & \textbf{1.420}                      & \textbf{0.477} & 1.187   \\\bottomrule
        \end{tabular}}
\end{table*}

\subsection{Structured Label Learning}
\label{sec:perception}

Text-based image understanding provides a flexible interface for interacting with VLMs.
However, many real-world applications require \emph{structured} visual perception outputs.
We focus on dense image captioning, which associates localized image regions with semantically rich captions.
For example, in autonomous driving, dense captioning can describe fine-grained scene elements (\eg, road type and location), while the model then performs higher-level reasoning over these elements.
We study two common formulations for learning bounding boxes under a next-token prediction objective.

\paragraph{Plain-Text Coordinates.}
A straightforward approach serializes a box as numeric coordinates, \eg, $[x_1, y_1, x_2, y_2]$, and trains the model to generate them as text~\cite{zhu2025internvl3,bai2025qwen2}.
While simple and compatible with standard LLM vocabularies, each coordinate may be split into multiple subword tokens, which can weaken the alignment between textual symbols and spatial locations.

\paragraph{Special Location Tokens.}
An alternative discretizes the image into a $K \times K$ grid and introduces dedicated location tokens $\{\text{loc}_{i}\}_{i=1}^{K}$~\cite{xiao2024florence,lu2024unified}.
A position is represented by predicting two tokens corresponding to its row and column indices.
This avoids fragmenting numeric coordinates into multiple tokens and provides a more explicit, semantically grounded representation of spatial positions. 

We conduct a controlled comparison of these two formulations on compact VLMs.
Our results (see \cref{sec:structured-label-details}) show that special location tokens consistently yield better dense captioning performance.
We hypothesize that discretized location tokens provide a clearer mapping between output symbols and image regions, which is particularly beneficial in the compact regime.
 \section{Experiments}\label{sec:exp}

This section describes our experimental setup and evaluates compact vision-language models (VLMs) on both text-based image understanding and dense image captioning (structured perception).

\paragraph{Datasets.}
We train \ours on a mixture of image-caption pairs, visual instruction data, and localization data with bounding-box annotations.
To support reproducibility, we rely exclusively on publicly available datasets.
For image captioning, we use 1.2M image-caption pairs from ShareGPT4V~\cite{chen2024sharegpt4v} and 591K pairs from the COCO2017 training split~\cite{lin2014microsoft}.
For instruction following, we combine 3.55M curated examples from ShareGPT4V~\cite{chen2024sharegpt4v}, WiT~\cite{srinivasan2021wit}, VFLAN~\cite{chen2024allava-vflan}, ScienceQA~\cite{lu2022scienceqa}, and MGM-Instruct~\cite{li2024mgm}.
To enable structured bounding-box prediction, we leverage box annotations from Objects365~\cite{shao2019objects365}, COCO2017~\cite{lin2014microsoft}, and Visual Genome~\cite{krishna2017visual}.

\paragraph{Evaluation Metrics.}
We evaluate \ours against leading compact VLMs on standard image understanding benchmarks:
VQAv2~\cite{goyal2017vqav2} for general visual question-answering, POPE~\cite{li2023pope} with a focus on object hallucination detection, and GQA~\cite{hudson2019gqa} for compositional reasoning and relational understanding.
For captioning, we report results on COCO2017~\cite{lin2014microsoft} and NoCaps~\cite{agrawal2019nocaps}.
For structured perception, we evaluate dense image captioning following the protocol of \citet{wu2024grit}.
We also measure inference efficiency via time-to-first-token (TTFT) and decoding throughput (tokens/second).

\begin{figure*}[t]
    \begin{subfigure}[b]{0.48\textwidth}
        \includegraphics[width=\textwidth]{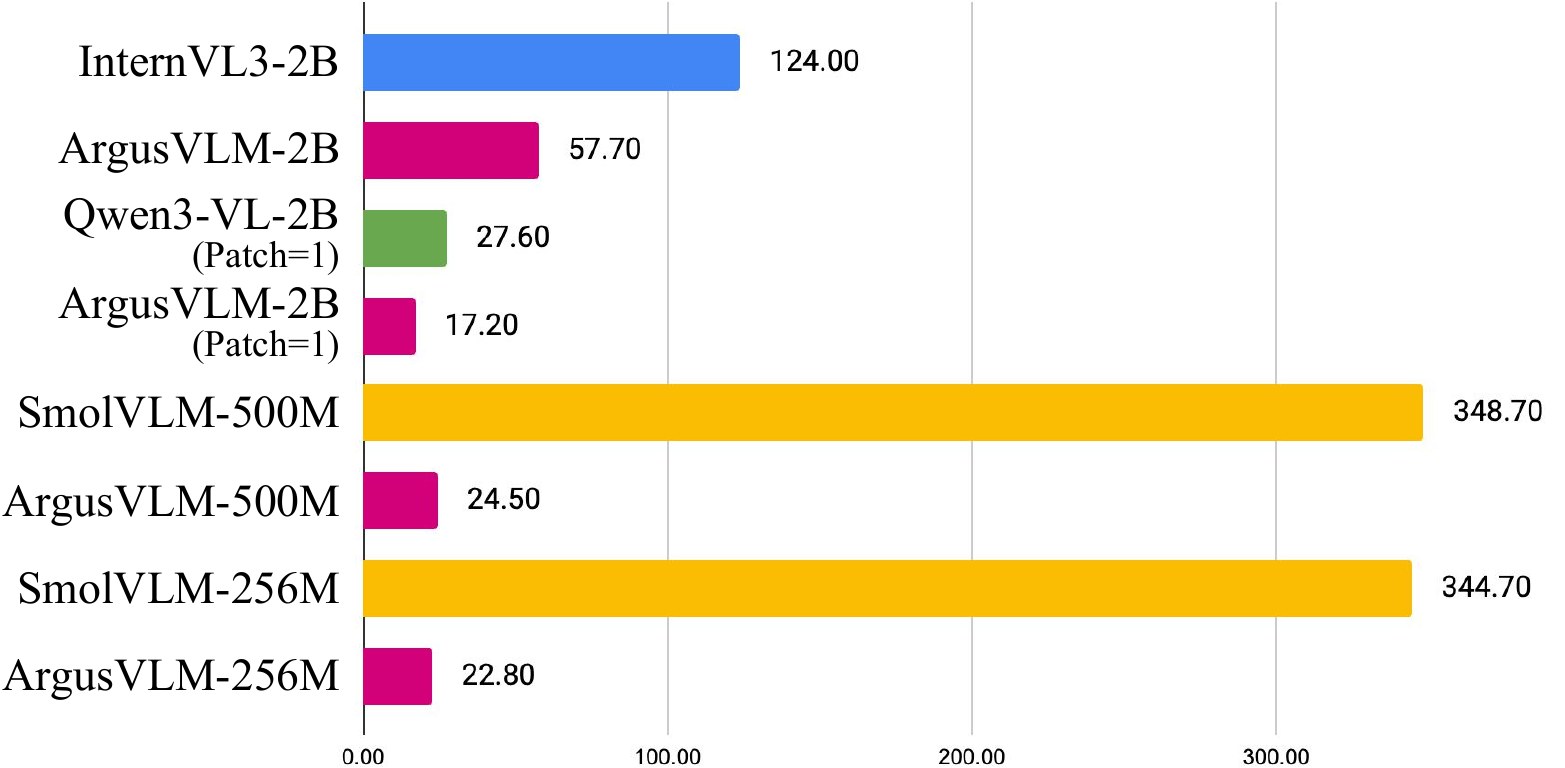}
        \caption{Time-to-first-token (ms) \(\downarrow\)}
        \label{fig:ttft}
    \end{subfigure}
    \hfill
    \begin{subfigure}[b]{0.48\textwidth}
        \includegraphics[width=\textwidth]{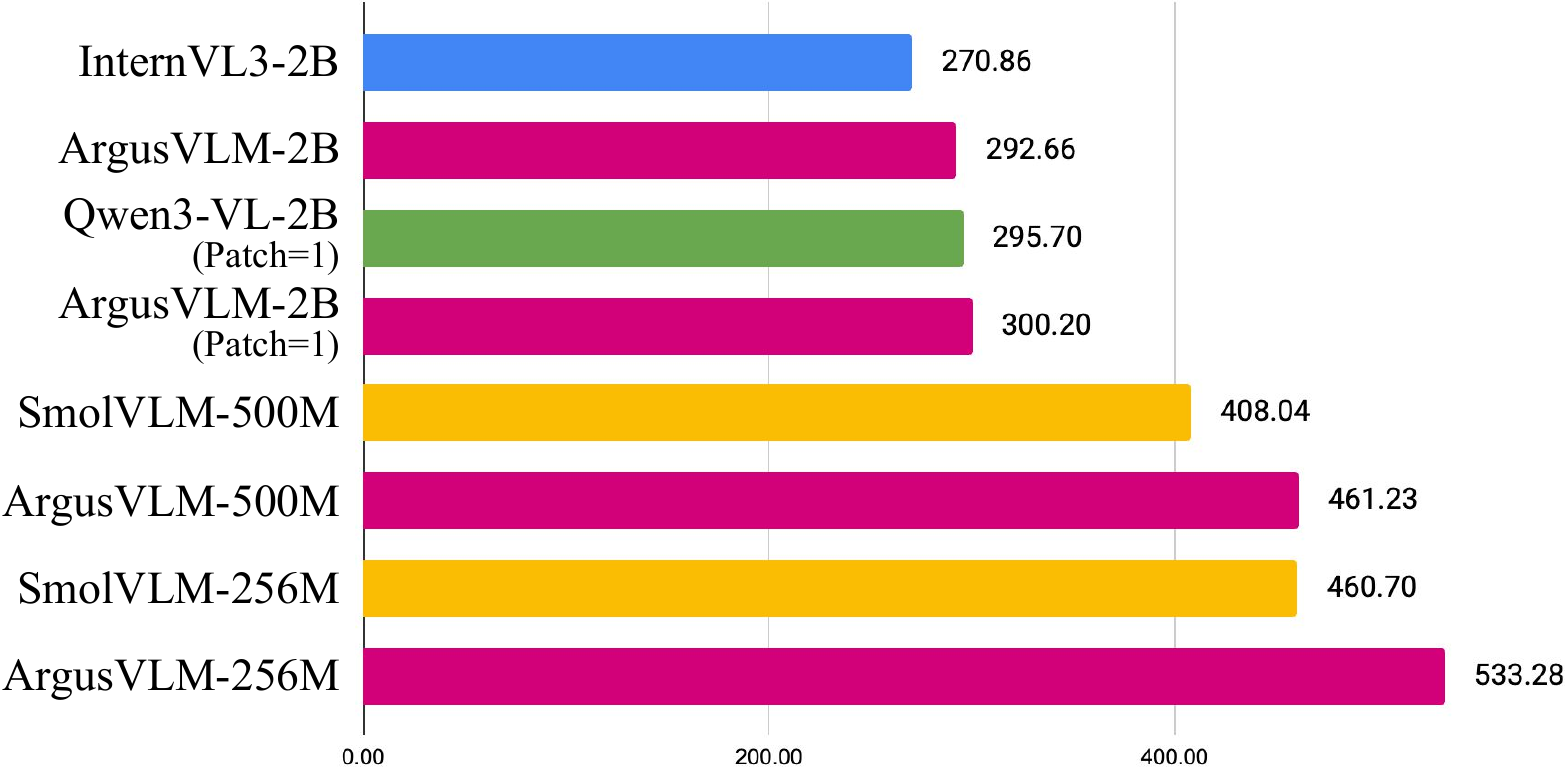}
        \caption{Decoding throughput (tokens/s) \(\uparrow\)}
        \label{fig:throughput}
    \end{subfigure}
    \caption{Efficiency comparison between \ours and leading VLMs.
        \ours outperforms both SmolVLM and InternVL3 on TTFT and throughput.
        As Qwen3-VL-2B does not use tiling, we compare it to \ours-2B under the patch=1 setting.
    }\label{fig:efficiency}
\end{figure*}

\paragraph{Implementation Details.}
We build \ours at three scales, from 256M to 2B parameters.
We bootstrap from publicly available pretrained checkpoints and apply supervised fine-tuning to both (i) improve efficiency and (ii) adapt the model to dense captioning.
Specifically, \ours-256M and \ours-500M are based on SmolVLM~\cite{marafioti2025smolvlm}, using a single-layer MLP to connect a SigLIP-B/16~\cite{zhai2023sigmoid} vision encoder (93M parameters) to SmolLM2-135M and SmolLM2-360M~\cite{allal2025smollm2} text decoders, respectively.
\ours-2B follows the same design but uses InternViT-300M~\cite{chen2024internvl} and Qwen2.5-1.5B~\cite{qwen2024qwen2_5} as the vision encoder and text decoder.
The input resolution is $512 \times 512$ for \ours-256M and \ours-500M, and $448 \times 448$ for \ours-2B.
After encoding image tiles and the thumbnail into visual embeddings, we apply pixel unshuffle~\cite{shi2016real} to reduce the number of visual tokens to 64 for \ours-256M/500M ($r=4$) and 256 for \ours-2B ($r=2$).
During inference, we apply the optimizations introduced in \cref{sec:case-study} to all model variants.

\begin{table}[t]
    \centering
    \small
    \caption{Comparison of bounding-box representations for dense captioning on Visual Genome.}
    \label{tab:densecap-ablation}
    \begin{tabular}{@{}lcc@{}}\toprule
        \textbf{Model}          & \textbf{Bbox Format} & \textbf{mAP (\%)} \\ \midrule
        \ours-256M              & Plain-text           & 1.89              \\
        \ours-256M              & Location Tokens      & 1.98              \\
        Florence2-base (0.23B)  & Location Tokens      & 1.67              \\
        \midrule
        \ours-500M              & Plain-text           & 2.18              \\
        \ours-500M              & Location Tokens      & 3.21              \\
        Florence2-large (0.77B) & Location Tokens      & 1.48              \\
        \bottomrule
    \end{tabular}
    \vspace{-1em}
\end{table}

\begin{figure*}[t]
    \centering
    \begin{subfigure}{0.97\textwidth}
        \includegraphics[width=\textwidth]{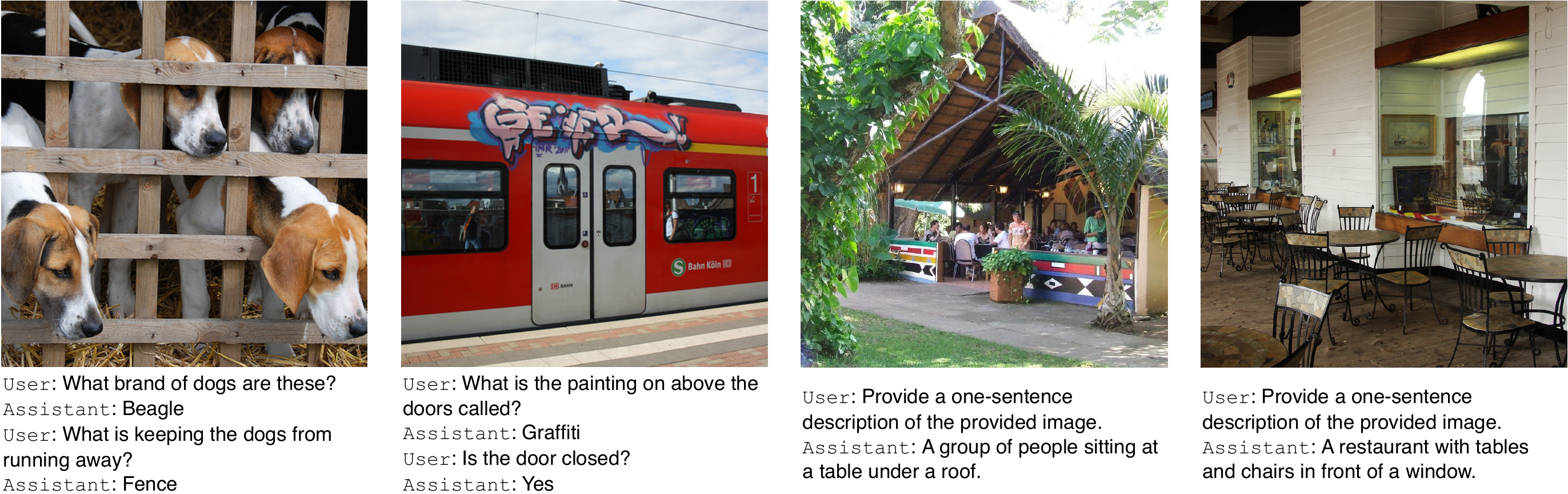}
\vspace{0.2em}
    \end{subfigure}
    \begin{subfigure}{0.97\textwidth}
        \includegraphics[width=\textwidth]{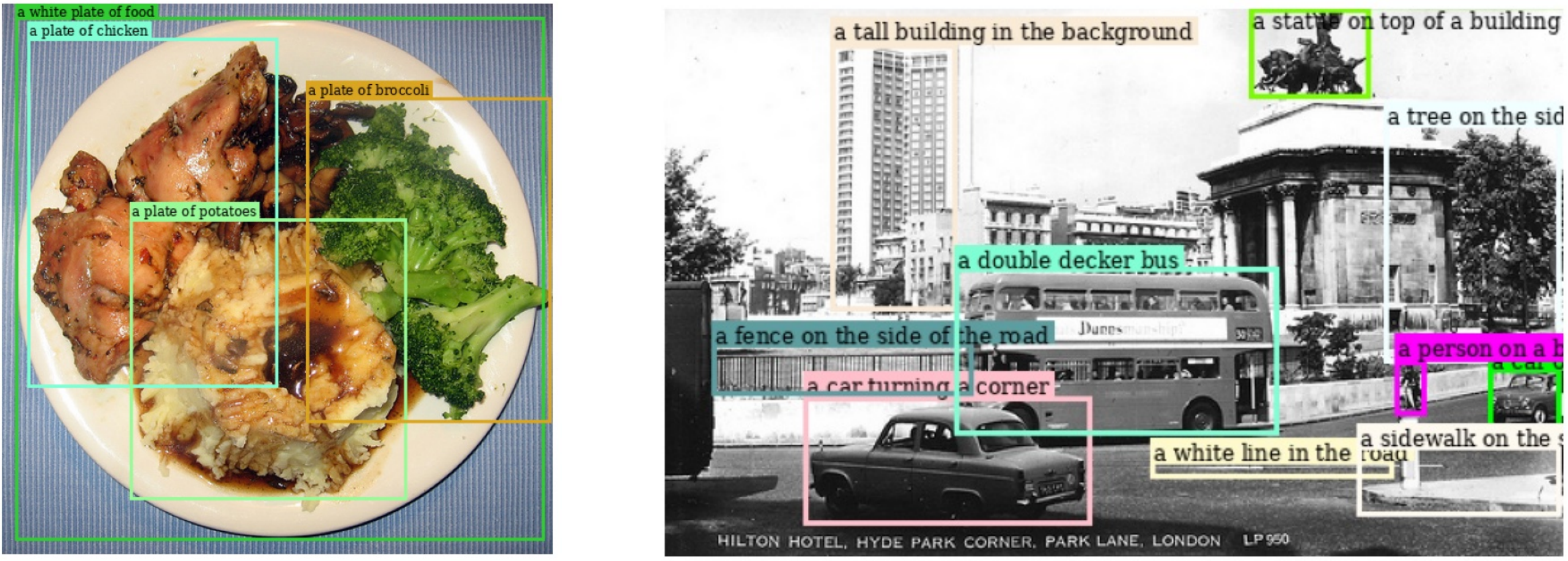}
\end{subfigure}
\caption{Qualitative results from \ours-2B. Top: VQA and image captioning. Bottom: dense captioning.}
    \label{fig:visual}
\end{figure*}

\subsection{Comparison on Visual Understanding}

As shown in \cref{tab:sota}, \ours achieves strong performance on diverse VQA and captioning benchmarks compared to existing compact VLMs with fewer than 2B parameters.
Compared to compact variants of recent state-of-the-art VLM families such as InternVL3~\cite{zhu2025internvl3} and Qwen3-VL-2B~\cite{bai2025qwen3vltechnicalreport}, \ours-2B achieves the best performance on 4 out of 5 image understanding tasks.
Notably, \ours-2B outperforms Qwen2.5-VL across all reported metrics with fewer parameters.
Restricting to models under 1B parameters,
\ours-256M and \ours-500M improve over SmolVLM counterparts by an average of 16\% and 11\% on the three VQA benchmarks.
Overall, \ours provides state-of-the-art accuracy in the sub-2B regime while remaining compact and efficient.

\subsection{Comparison on Dense Captioning}\label{sec:structured-label-details}

We evaluate dense image captioning on Visual Genome~\cite{krishna2017visual}, which requires predicting a set of regions together with free-form captions describing each region.
Following prior work~\cite{johnson2016densecap,xiao2024florence}, we compute a joint mean Average Precision (mAP) over region--caption pairs by treating each paired prediction as a retrieval target.
We compare two ways of representing bounding boxes within the language modeling space: (i) plain-text coordinates and (ii) special location tokens.
Both keep the rest of the output in natural language, maintaining compatibility with standard next-token prediction.
We evaluate \ours at the 256M and 500M scales and compare against Florence2-base and Florence2-large~\cite{xiao2024florence}, which also use location tokens.

As shown in \cref{tab:densecap-ablation}, special location tokens consistently outperform plain-text coordinates for all \ours variants.
We attribute the gap primarily to reduced token fragmentation: location tokens replace multi-token numeric substrings with single, semantically homogeneous units.
This simplifies optimization during fine-tuning, reduces exposure bias from partially generated numbers, and helps the model learn stable joint distributions over \{position, caption\}.
The gains increase with model scale, suggesting that larger models benefit more from discrete spatial representations, while the 256M model remains capacity-limited.

\subsection{Comparison on Inference Efficiency}\label{sec:ablation-efficiency}

We compare \ours with InternVL3-2B~\cite{zhu2025internvl3}, Qwen3-VL-2B~\cite{bai2025qwen3vltechnicalreport}, and multiple SmolVLM variants~\cite{marafioti2025smolvlm} in \cref{fig:efficiency}.
For fair comparison, we use batch size 1, run all methods on an NVIDIA H100 GPU, and average over the same COCO2017 image-captioning requests.
Baselines use their official vLLM implementations without applying our optimizations (\cref{sec:method-efficiency}).
With our proposed optimizations, \ours achieves the lowest TTFT and among the best decoding throughput across model sizes.

\subsection{Applying the Optimization Recipes Beyond vLLM}\label{sec:more-than-vllm}

The optimization recipes in \cref{sec:recipes} also apply to serving stacks beyond vLLM.
Following the same iterative profiling workflow, we optimized the Hugging Face~\cite{wolf2019huggingface} implementations of InternVL3-2B and SmolVLM-256M.
Optimizing image processing and text tokenization reduces TTFT by 28.3\% (145\,$\rightarrow$\,104\,ms) for InternVL3-2B and by 63.3\% (332\,$\rightarrow$\,122\,ms) for SmolVLM-256M.
We also identified and improved slow GPU-side components (notably RMSNorm and rotary positional embeddings) and removed redundant operations.
These changes improve throughput by 47.5\% (51.6\,$\rightarrow$\,98.2 tokens/s) and 43.3\% (53.2\,$\rightarrow$\,93.9 tokens/s), respectively.

\subsection{Visualization}

\cref{fig:visual} provides qualitative examples of \ours-2B.
The examples in the top row show visual question answering and captioning predictions across diverse scenes.
\ours correctly answers user queries, and produces coherent captions conditioned on the images.
The bottom row shows \ours's capability on predicting bounding boxes along with description for dense captioning.
These examples demonstrate that \ours supports structured perception outputs while retaining the flexible natural-language interface of text-based image understanding.
 \section{Conclusion}
We introduced \ours, a compact and efficient vision-language model series that unify text-based visual understanding with fine-grained perception.
Through end-to-end profiling of compact VLM inference, we identified key implementation bottlenecks and proposed targeted optimizations that significantly reduce latency without degrading accuracy.
To extend compact VLMs to structured perception, we further studied effective formulations for learning spatially grounded outputs (\eg, bounding boxes) under a next-token prediction objective.
Across extensive experiments, \ours delivers strong performance while remaining compact, and achieves substantially faster inference than existing compact VLMs.
Overall, our results provide practical guidance and empirical evidence for building efficient and capable vision-language systems.

\clearpage

\bibliography{main}
\bibliographystyle{icml2026}

\end{document}